\documentclass[letterpaper, 10 pt, conference]{ieeeconf}  

\IEEEoverridecommandlockouts                              

\usepackage{xcolor}
\usepackage{graphics} 
\usepackage{amsmath} 
\usepackage{amssymb}  
\usepackage{graphicx} 
\usepackage{adjustbox}
\usepackage[ruled, noend]{algorithm2e}
\SetKwInput{kwInit}{Init}
\SetKwComment{Comment}{$\triangleright$\ }{}
\SetCommentSty{itshape}
\DeclareMathOperator*{\argmax}{arg\,max}
\usepackage{todonotes}  
\usepackage{booktabs} 
\usepackage{siunitx} 
\usepackage{multirow}
\usepackage{soul} 
\usepackage{makecell} 

\graphicspath{{figures/}}
\usepackage{subcaption}
\title{\LARGE \bf
Learning to Arbitrate Human and Robot Control \\
using Disagreement between Sub-Policies}

\author{Yoojin Oh$^{1}$\thanks{Corresponding author: \tt{yoojin.oh@ipvs.uni-stuttgart.de}}, Marc Toussaint$^{2,3}$ and Jim Mainprice$^{1,3}$\\
\authorblockA{$^1$Machine Learning and Robotics Lab, University of Stuttgart, Germany}
\authorblockA{$^2$Learning and Intelligent Systems Lab ;  TU Berlin ; Berlin, Germany}
\authorblockA{$^3$Max Planck Institute for Intelligent Systems ;  MPI-IS ; T{\"u}bingen/Stuttgart, Germany}
\vspace{-0.8cm}
}

\begin{document}
\bstctlcite{IEEEexample:BSTcontrol}

\maketitle
\thispagestyle{empty}
\pagestyle{empty}

\begin{abstract}

  In the context of teleoperation, 
  arbitration refers to deciding how to blend between human and autonomous robot commands. 
  We present a reinforcement learning solution that learns an optimal arbitration strategy 
  that allocates more control authority to the human 
  when the robot comes across a decision point in the task. 
  A decision point is where the robot encounters multiple options (sub-policies), 
  such as having multiple paths to get around an obstacle or deciding between two candidate goals. 
  By expressing each directional sub-policy as a von Mises distribution, 
  we identify the decision points by observing the modality of the mixture distribution. 
  Our reward function reasons on this modality  
  and prioritizes to match its learned policy to either the user or the robot accordingly.
  We report teleoperation experiments on reach-and-grasping objects using a robot manipulator arm 
  with different simulated human controllers. 
  Results indicate that our shared control agent outperforms direct control 
  and improves the teleoperation performance among different users. 
  Using our reward term enables flexible blending between human and robot commands 
  while maintaining safe and accurate teleoperation.

\end{abstract}

\section{INTRODUCTION}

The level of autonomy in robot teleoperation,
and hence shared control
between the human and the robot, 
is a hot topic of both academic
	and industrial research
\& development.
Shared autonomy is especially
desirable when the environment is structured but
the task objectives are unspecified and have to be
decided in real-time, such as disaster relief \cite{phillips2016autonomy}, 
or autonomous driving \cite{johns2016exploring}, or 
the control of assistive devices, for instance, 
controlling a robotic arm using a cortical implant
\cite{muelling2017autonomy} or a robotic wheelchair,
where the human control authority
is a key feature \cite{goil2013using}.

\begin{figure}[t]
\begin{center}
  \hspace{-0.3cm}
  \includegraphics[width=1\linewidth,clip,trim={2.5cm 1.5cm 4cm 1.5cm}]{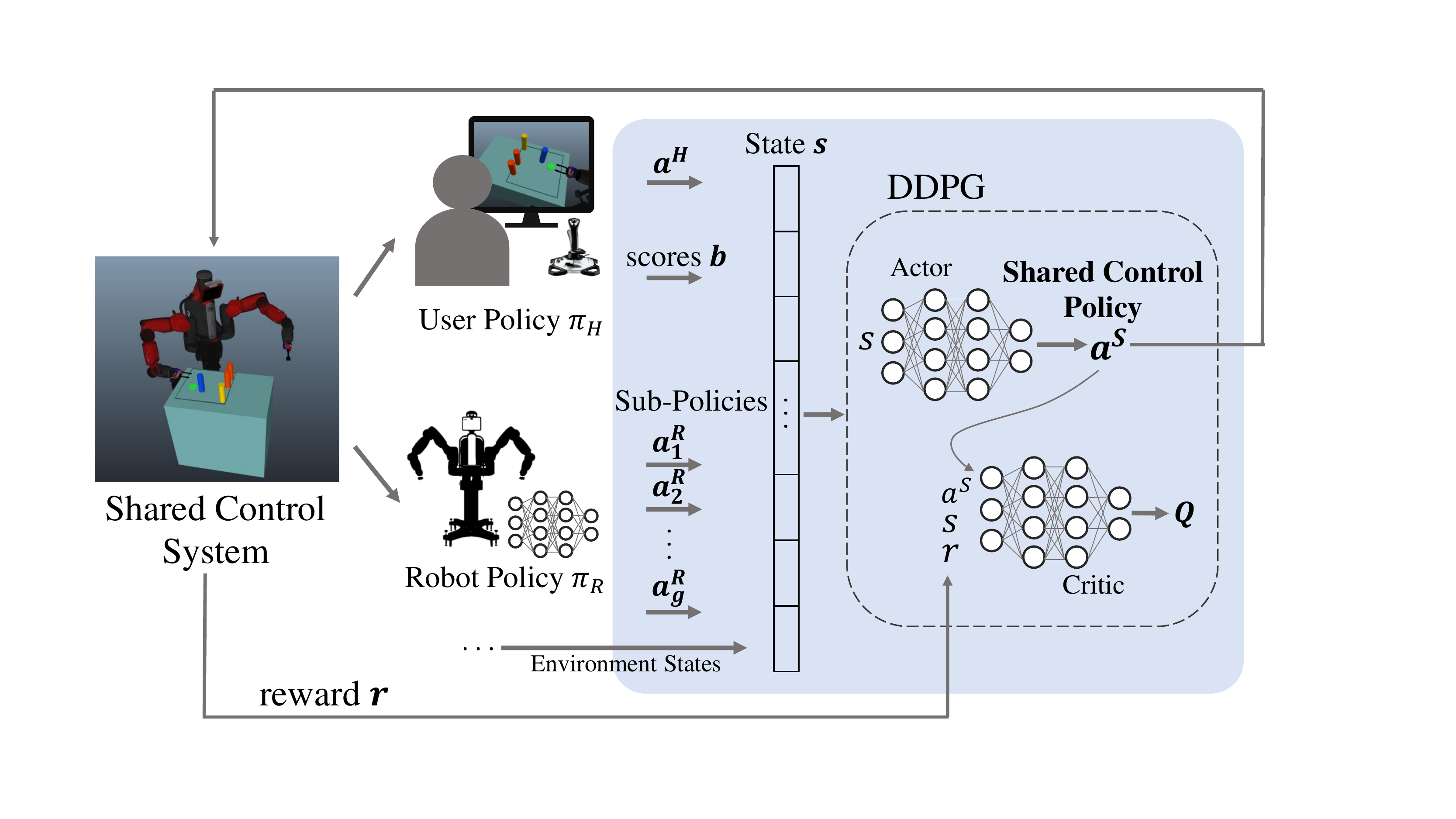}
  \vspace{-.3cm}
  \caption{Overview of our method using the DDPG (Deep Deterministic Policy Gradient) algorithm. 
  The arbitration module takes as input the user policy, the sub-policies for all goals, their scores, 
  and outputs a continuous arbitrated policy. }
  \label{fig:workflow}
 \vspace{-.7cm}
\end{center}
\end{figure}

{

Traditional approaches to robot teleoperation
rely on the user assigning low-level (\textit{direct control}) or mid-level (\textit{traded control})
commands to be performed by the robot.
In these cases, the human always has control over the robot's actions.
In contrast, \textit{shared control} arbitrates between
human and robot control by inferring the human intent,
and executing the action autonomously when possible.
Shared control can mitigate difficulties linked to direct control,
which originate from human bounded rationality,
limited situation awareness
and the discrepancy between the human and robot morphologies.

Shared autonomy systems rely on two components:
\begin{enumerate}
\item prediction of user intent
\item blending user vs. autonomous controls
\end{enumerate}

Trading off robot and human control 
is usually directly based on the confidence scores
associated with the intent prediction module.
This paper focuses on the blending strategy (i.e., arbitration),
and leaves the prediction to other works.



Our approach (see Figure \ref{fig:workflow}) learns an arbitration strategy from user interaction.
This is generally beneficial as different users may have
different preferences.
Though tempting, it is difficult to learn an arbitration
function using supervised learning.
Indeed the introduction of the arbitration will inevitably
shift the data distribution and hence the
behavior of the human in return.

Several works have studied the problem of using
reinforcement learning in shared control.
However, to the best of our knowledge, no work
has focused on learning directly the arbitration module
while leveraging the local geometry of the available sub-policies.
This is desirable to limit the sample complexity by
1) limiting the search space by using readily available optimal actions, 
2) making use of the disagreement between the sub-policies and the shared actions to create implicit feedback, 
resulting in dense rewards.

Finally, to solve continuous action spaces control
tasks common in robotic applications, 
we apply the modern actor-critic algorithm Deep Deterministic Policy Gradient (DDPG) \cite{lillicrap2015continuous}.

We summarize our main contributions as the following:
\begin{itemize}
	\item A generic arbitration learning formulation,
	 which we cast as an actor-critic reinforcement learning problem
	 
  \item Identification of ``decision points'' in the environment, 
  where the robot encounters multiple options

	\item A new reward term that allows to maximize human control authority 
	near decision points, by implicit feedback in continuous domains

	\item Quantitative results in simulation that show
	the effectiveness of the approach
	on a realistic robot manipulation scenario.
	
\end{itemize}

This paper is structured as follows: 
We first go over related work in Section~\ref{sec:related-work},
we then present our method in Section~\ref{sec:methods}. 
We introduce our framework and explain the implementation in Section~\ref{sec:experiment_setup}, 
and results are discussed in Section~\ref{sec:results}. 
Finally, conclusions are drawn in Section~\ref{sec:conclusions}. 

\section{Related Work}
\label{sec:related-work}

\subsection{Arbitration in Shared Control}

One common form of blending is through a linear combination between the human user and autonomous robot agent policies
\cite{dragan2013policy, allaban2018blended, schultz2017goal, gopinath2016human}.
Generally the arbitration function $\alpha : ( a^H, a^R ) \mapsto a^S$, 
given user and robot actions $a^H, a^R$ and outputs an arbitrated action $a^S$, 
can depend on different factors such as the confidence in the user intent prediction
\cite{dragan2013policy, gopinath2016human, schultz2017goal},
or considering the difference between each commands
\cite{allaban2018blended}.
}

{

When the robot predicts the user's intent with high confidence,
the user often loses \textit{control authority}.
This has been reported to generate mixed preferences from users
where some users prefer to keep control authority despite longer completion times~\cite{kim2011autonomy, Javdani:2018bt}.
Additionally, when assistance is against the user's intentions
this approach can aggravate the user's workload~\cite{dragan2013policy};
the user ``fights" against the assistance rather than gain help from it.
Defining an arbitration function that is nor too 
\textit{timid} (i.e., only gives assistance when very confident)
nor to \textit{aggressive}, is generally difficult
and interpreting the noisy confidence estimate of the
intent prediction is prone to errors.
}

{

Some works have taken the option to
allocate maximum control authority to the user
by minimally applying assistance only when it is necessary.
Broad et. al.~\cite{broad2018operation,broad2019highly} introduced
minimum intervention shared control that computes whether the control signal 
leads to an unsafe state and replaces the user control if so.
However, these works usually model the problem by selecting
an action within a feasibility set. Here we
tackle problems with continuous action where
such a set is difficult to characterize.
}

{

To alleviate this issue we have proposed
``Natural gradient shared control" \cite{0h:20}, 
where we formulate shared control as an optimization problem. 
The shared action is chosen to maximize
the user's internal action-value function while constraining the
shared control policy to not diverge away from the autonomous robot policy.
This approach is based on the availability of
a global goal-conditioned optimal policy for the robot.
This is nowadays only available for low-dimensional
domains with perfect state estimation, however it is
anticipated that with the advances of reinforcement learning
\cite{mnih2015human, silver2016mastering}, such control policies
should become available for an increasing number of domains.
However, despite this formulation of the problem, it is desirable
to tune the arbitration online so as to account for user preferences.
}

{


\subsection{Reinforcement Learning for Shared Control}


Recent works have proposed to use reinforcement learning
for shared control (RLSC) in different application contexts.
In \cite{xu2015reinforcement}, SARSA, an on-policy reinforcement learning algorithm, 
is used to produce an arbitration weight for a walking-aid robot.
The reward function penalizes collision with environment 
and promotes smoothness.
In contrast to our work, no \textit{implicit feedback} from the user is considered.

The notion of implicit feedback as a reward term is introduced 
in \cite{gaoxt2} to train an agent for a X-to-Text application.
The error correction input (i.e., backspace) from the user is used to penalize
the wrong actions taken by the autonomous agent.

In recent years, some model-free RLSC methods
have been proposed.
The earliest work is \cite{reddy2018shared}, 
where the agent maximizes a combination of task performance and user feedback
rewards using Deep-Q learning.
In \cite{schaff2020residual}, the approach is extended
to maximize human-control authority
using residual policy learning.
While general, model-free approaches do not
consider the structure of the problem to maximize adaptability
at the cost of sample complexity.
Instead, our approach investigates the case where robot
sub-policies are available. In our experiments, these correspond
to goal-conditioned policies, but they do not have to be.

In \cite{fernandez2021deep}, Fernandez and Caarls develop
an RLSC agent to learn a haptic policy using a type of implicit feedback,
which compares the velocity the user applies with the forces exerted.
While similar to our approach, their approach does not make use of an available optimal policy,
which is instead learned online.

The ideas developed in RLSC have been generalized in \cite{du2020ave},
which formalizes a framework for assistive systems where the notion of
human \textit{empowerment} is introduced to denote reward terms
that promote the user control over the state.
The reward function we propose can be viewed as a type of empowerment proxy,
especially designed to account for sub-policies in continuous action spaces,
which are common in robot control problems.

}

\section{Reinforcement Learning for Arbitration}
\label{sec:methods}
\subsection{Problem Setting}

In shared control where a human agent and the robot agent shares control to accomplish a mutual task, the state space includes the user action $a^{H} \in \mathcal{A}^{H}$, the robot actions $a^{R}_{g} \in \mathcal{A}^{R}_{g}$ for each goal $g \in \mathcal{G}$. 
Each robot action $a^{R}_{g}$
may result from a different sub-policy indexed by $g$,
in our experiments
we simply conditioned a single policy on $g$.
Each goal or sub-policy is associated with
a confidence score or belief $b_g$.
The shared control agent produces an arbitrated action $a^{S} \in \mathcal{A}^{S}$.

\subsection{Learning an Arbitrated Policy}

A reinforcement learning agent for arbitration
strategy, learns a policy $\mu ( a_t^S | x_t) $, which takes as input
$$
s_t = \{x, a^{H}, a^{R}_{1}, \ldots, a^{R}_{g},  b_{1} , \ldots, b_{g} \}
$$
\noindent
where $x$ is the environment states (e.g. gripper position, distance to goals, distance to obstacle).
Our reinforcement learning problem is modeled
as an optimal control task of a Markov Decision Process (MDP),
which is defined by a set of states $\mathcal{S}$, a set of actions $\mathcal{A}$, transition transition probabilities $\mathcal{T}:\mathcal{S} \times \mathcal{A} \times \mathcal{S}$, a reward function $\mathcal{R}:\mathcal{S} \times \mathcal{A} \rightarrow \mathbb{R}$, and a discount factor $\gamma \in [0,1]$. 

To devise the arbitration strategy $\mu$, we follow the DDPG algorithm~\cite{lillicrap2015continuous}, but we point out two adjustments that distinguish our method from the standard version, both are reported in Algorithm \ref{alg:ddpg}.

\paragraph{Sub-policies}
We utilize all sub-policies when considering the arbitration rather than one policy. 
At each time step of the episode, we query all possible sub-policies towards each object and compute the scores.

In our experiments, the sub-policies $a^{R}_{t,g}$ represent the action for each prospective goal object $g$ and 
details of the sub-policies are later explained in Section~\ref{subsection:Subpolicies}. 
The score $b_g$ denotes how likely the object is the goal and it can be described as the posterior probability 
given a history of observed features $\xi_{S\rightarrow U}$~\cite{dragan2013policy, jain2019probabilistic}. 
The predicted goal object $g^*$ naturally becomes the object with the highest score. 
However, we do not emphasize one goal when learning the arbitrated action. 
We let the network take into account the inaccuracy of the intent prediction.

\paragraph{Hindsight goal labeling} Since we do not explicitly provide the true goal $g^*$ that the user is intending, 
we store the state-action pairs in a separate episode buffer $R_E$ 
and compute the rewards for each state-action pair in hindsight once the episode terminates. 
The off-policy characteristic of the DDPG algorithm enables us to update the replay buffer $R$ in a delayed manner,
as it already randomly samples a minibatch from $R$ to update the parameters of the network. 
We first collect $n$ episodes to fill the replay buffer to compensate the delayed population of $R$.

We propose an arbitration strategy that allocates more control authority 
when the robot needs to decide. 
Rather than making imperfect intent predictions to decide on which goal or direction to assist towards, 
we actively let the user decide instead. 
These decision points are identified by observing how the sub-policies diverge 
and creating implicit feedback in the form of a specific reward term.

\begin{algorithm}[ht]
\DontPrintSemicolon
\SetAlgoNoLine
 \kwInit{\\
 Load actor $\mu(s|\theta^{\mu})$, critic $Q(s,a|\theta^{Q})$ weights $\theta^\mu$, $\theta^{Q}$ \\
 Initialize target network $\mu'$, $Q'$ with weights $\theta^{\mu'}\leftarrow \theta^{\mu}$,$\theta^{Q'}\leftarrow \theta^{Q}$\\
 Initialize replay buffer $R$, episode buffer $R_E$
 }
 \For{$\textnormal{episode}=1, M$}{ 
 \For{$t=1,T$}{
  Observe user action $a^H_t$\;
  \ForEach{$g \in \mathcal{G}$}{
    Query sub-policy action $a^{R}_{t,g}$\;
    Compute score $b_{t,g}$\;
  }
  Select action $a^S_t=\mu(s_t|\theta^{\mu})+\mathcal{N}_t$\;
  Execute $a^S_t$, observe next state $s_{t+1}$, done {$d_t$}\;
  Store transition ($s_t, a^{S}_t, s_{t+1}, d_t$) in $R_E$\;
  \If{$t>n$}{
  Sample a random minibatch of $N$ transitions ($s_i, a_i, r_i, s_{i+1}, d_t$) from $R$\;
  Update $\theta^{\mu}, \theta^{\mu'}$, $\theta^{Q}, \theta^{Q'}$\\\;
  }
 }
 Observe true goal $g^{*}$\;
 \ForEach{$o \in R_E$}{
 Compute reward $r_i=\textit{Reward}(o,g^{*})$\; 
 Store transition ($s_i, a^{S}_i, r_i, s_{i+1}, d_t$) in $R$\;
 }
 }
 \caption{DDPG for Arbitration Learning}
 \label{alg:ddpg}
\end{algorithm}
\setlength{\textfloatsep}{1pt}

\begin{figure*}[h!]
\begin{minipage}[b]{0.17\linewidth}
\begin{subfigure}{1\textwidth}
\includegraphics[width=\linewidth,clip,trim={0.5cm 0cm 14cm 0cm}]{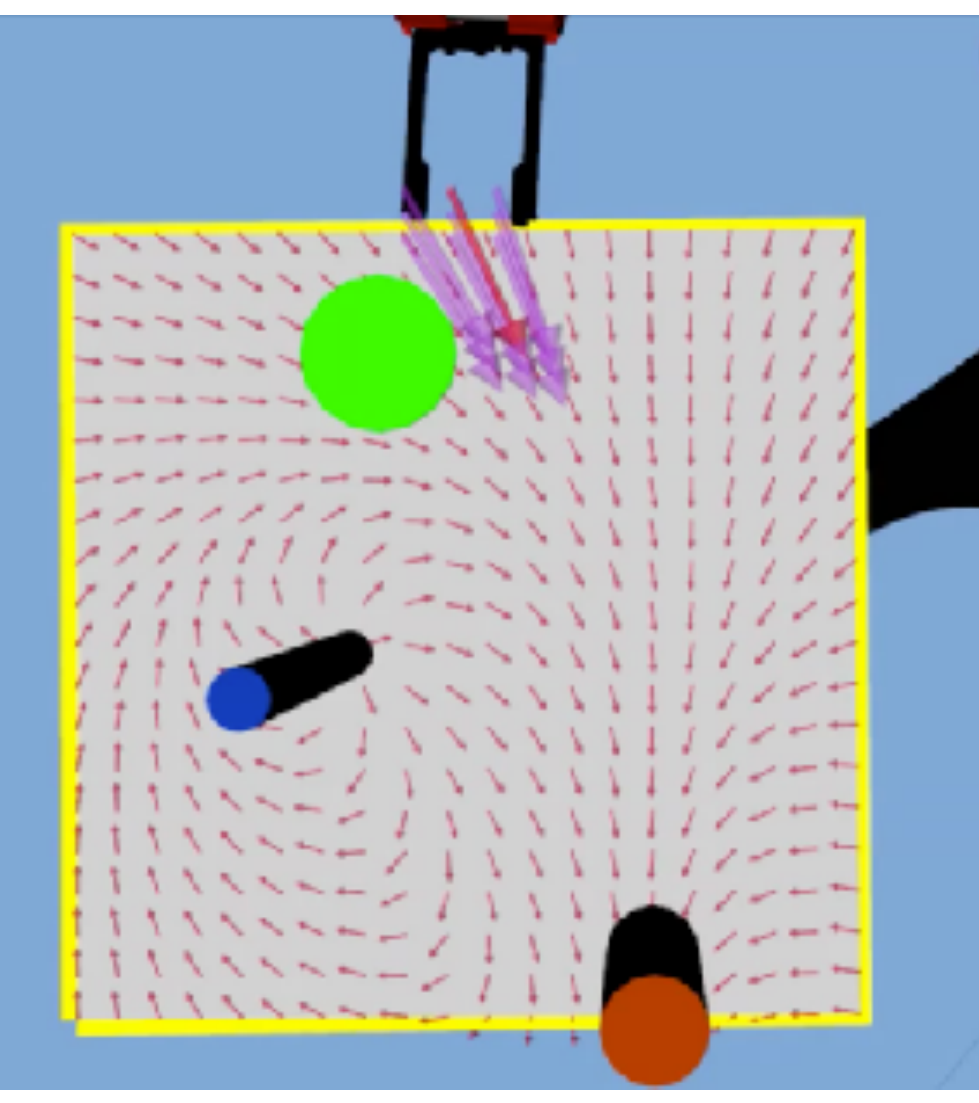}
\caption{Sub-policy for a goal object}
\end{subfigure}

\begin{subfigure}{1\textwidth}
\includegraphics[width=\linewidth,clip,trim={7cm 0cm 7.5cm 0cm}]{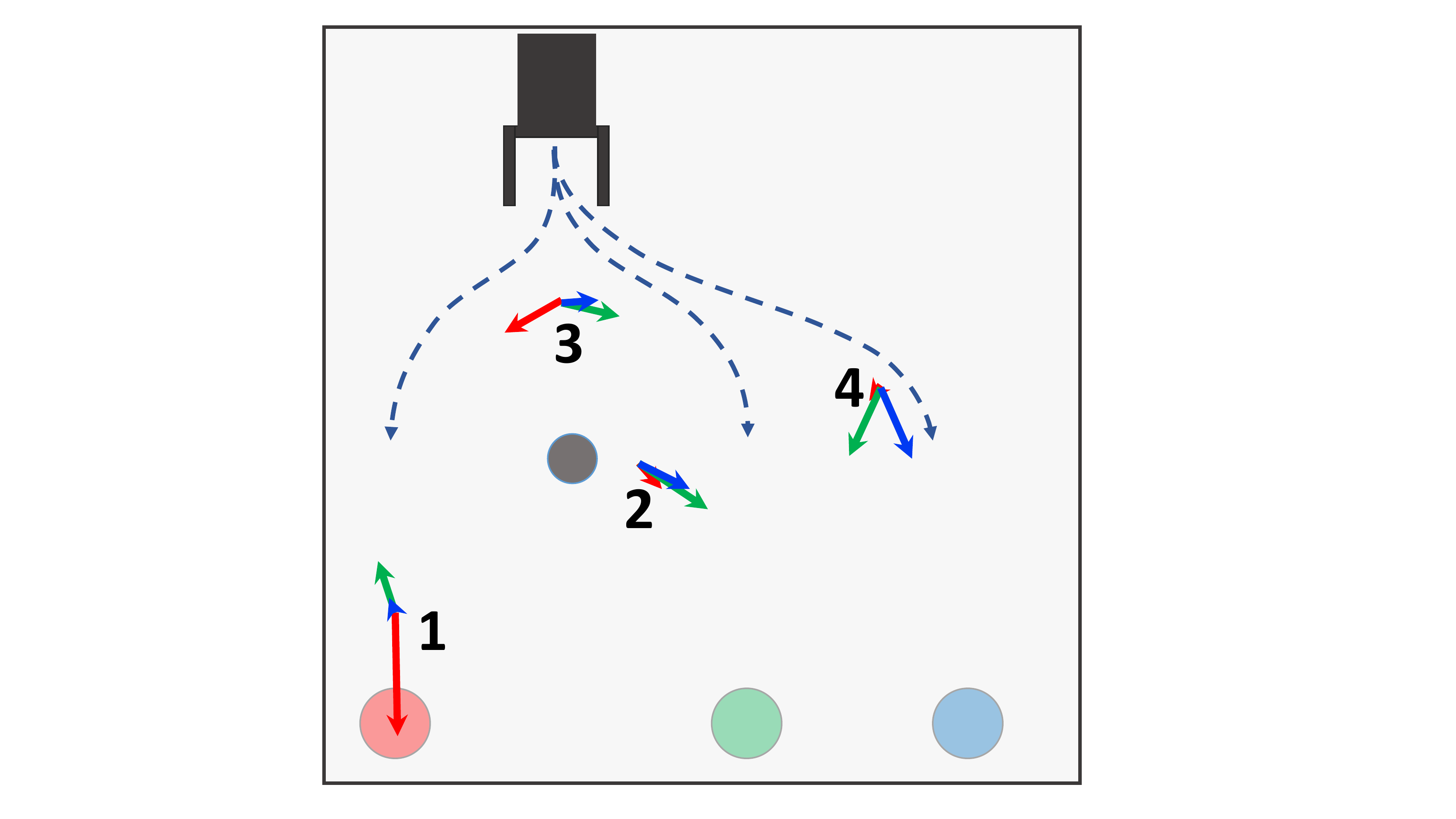}
\caption{Sub-policies at sampled positions}
\label{subfig-2:env}

\end{subfigure}
\end{minipage}
\begin{minipage}[b]{0.77\linewidth}
\begin{subfigure}{0.55\linewidth}
  \centering
  \includegraphics[width=\linewidth,clip,trim={2.2cm 1.7cm 3cm 3.7cm}]{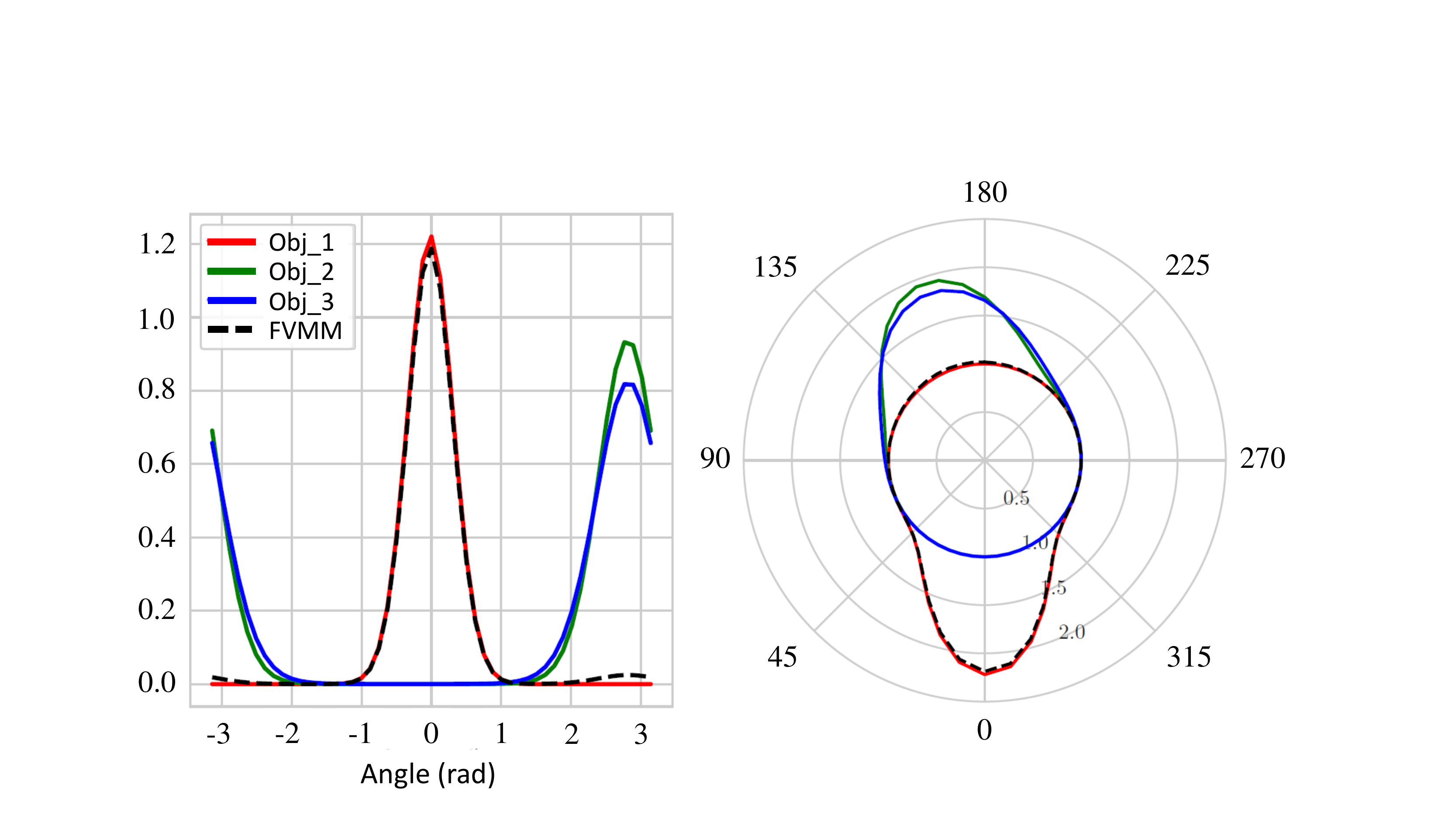}
  \caption{Point 1, Scores: 0.973, 0.021, 0.006}
  \label{subfig-2:p21}
  \end{subfigure}
  \begin{subfigure}{0.55\linewidth}
  \centering
  \includegraphics[width=\linewidth,clip,trim={2.2cm 1.7cm 3cm 3.7cm}]{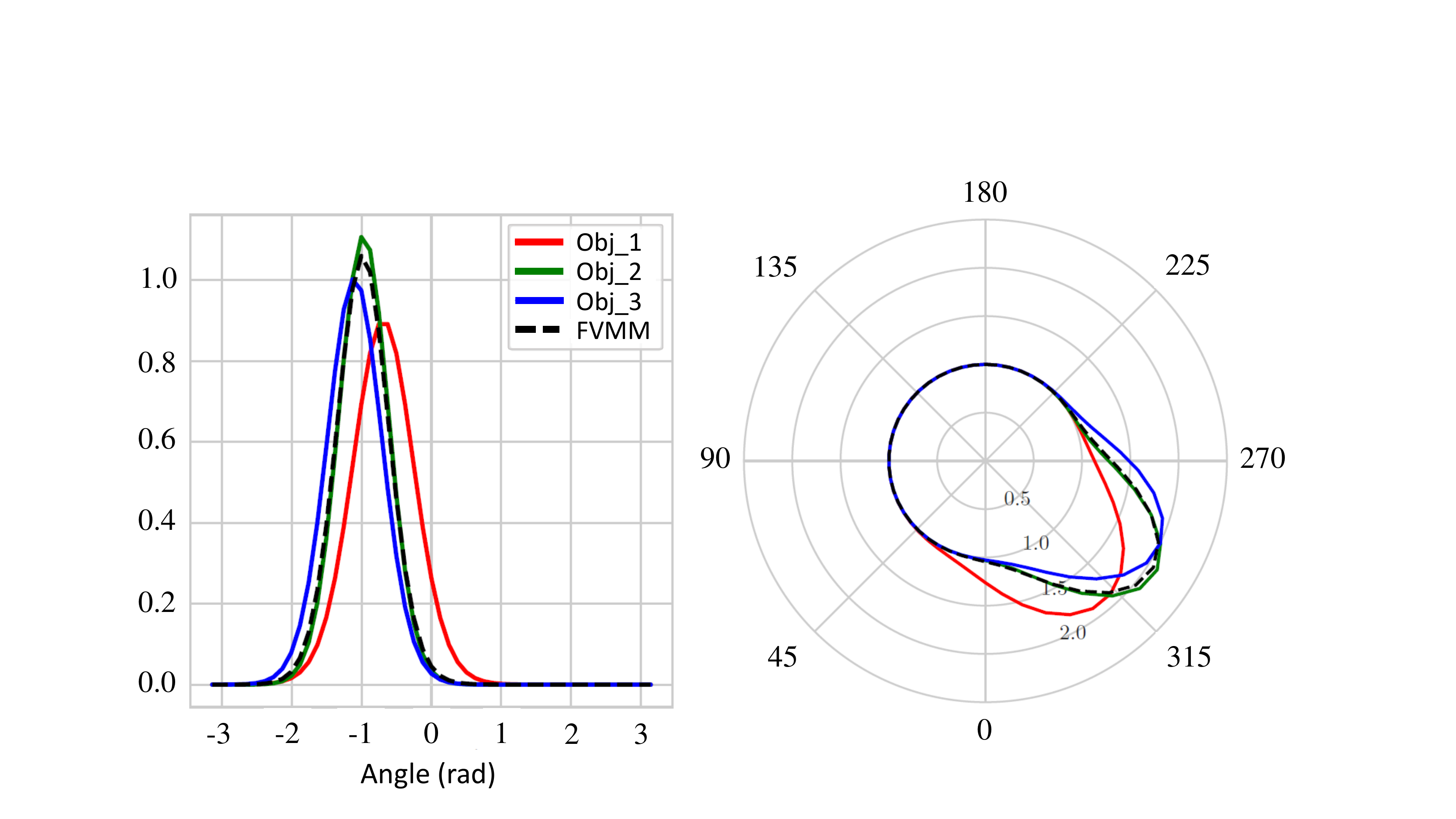}
  \caption{Point 2, Scores: 0.054, 0.761, 0.185}
  \label{subfig-2:p44}
  \end{subfigure}
  
  \begin{subfigure}{0.55\linewidth}  
  \centering  
  \includegraphics[width=\linewidth,clip,trim={2.2cm 1.7cm 3cm 3.7cm}]{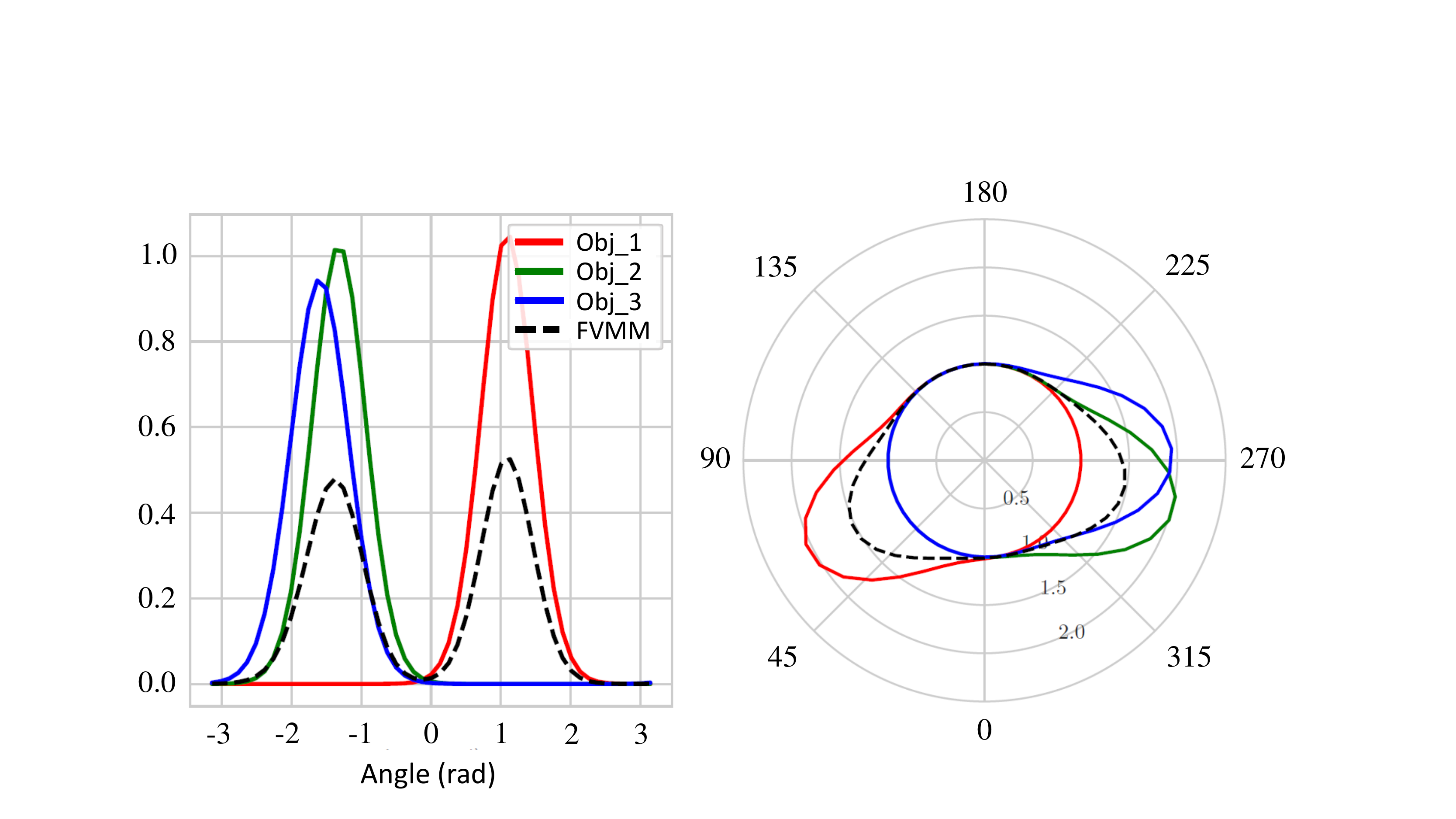}
  \caption{Point 3, Scores: 0.031, 0.445, 0.524}
  \label{subfig-3:p57}
  \end{subfigure}
  \begin{subfigure}{0.55\linewidth}
  \centering
  \includegraphics[width=\linewidth,clip,trim={2.2cm 1.7cm 3cm 3.7cm}]{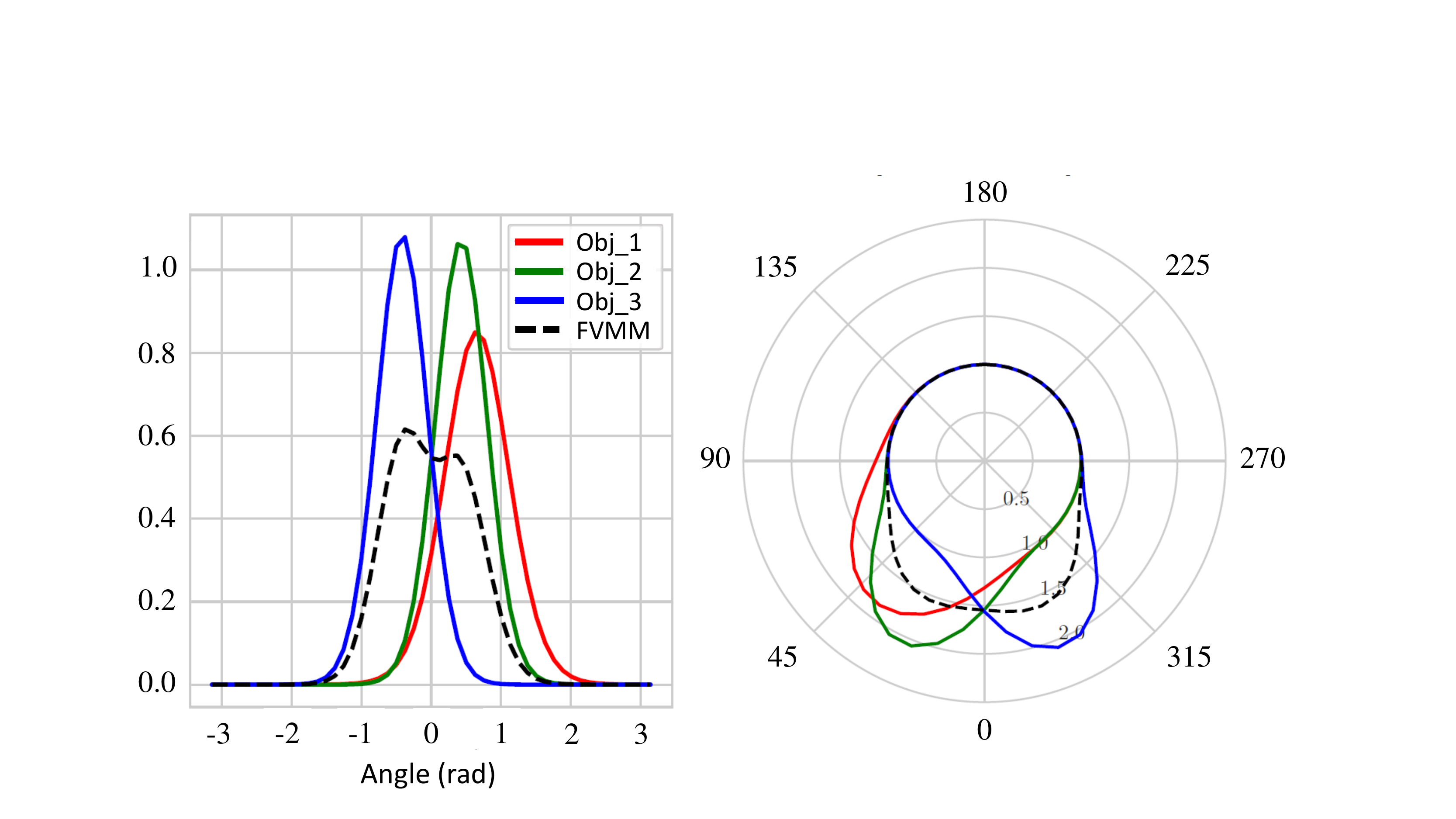}
  \caption{Point 4, Scores: 0.503, 0.362, 0.135}
  \label{subfig-3:p63}
  \end{subfigure}
\end{minipage}
    \caption{
    Von Mises distributions for each sub-policy (solid colored line, matching the colors of the objects in (b)) 
    and its Finite von Mises Mixuture Model (FVMM) (black dashed line). 
    Top row: Conditions when the robot should take control, 
    clear intention towards an object (c), similar sub-policies (d). 
    Bottom row: Conditions when the human should take control, 
    deciding which direction to get around the obstacle (e), in between two perspective goal objects (f).}
    \label{fig:mixture_distributions}
\vspace{-0.2cm}
\end{figure*}

\subsection{Identifying Decision Points}
A decision point is where one has to choose an option given multiple options. 
A robot agent may encounter them explicitly through intersections or getting around obstacles, 
or implicitly such as deciding to choose a goal given multiple prospective goals..
In a human-robot collaboration task such as teleoperation, inferring the user intent or the goal can mitigate this problem by estimating the belief over the possible goals. 
However, the intent may only be more certain once the robot has passed the decision point. 
 
Our approach lets the user take more control when there is a decision point in the environment. 
In a problem setting with a discrete number of goals, 
we identify a decision point by looking at how the sub-polices for the goals are distributed. 
When the robot encounters an obstacle, the sub-policies show two divergent paths that the robot can take. 
Similarly, when the robot is between two goal objects, the policies for each object will result in two divergent paths that the robot can take. 
Rather than relying on an intent prediction method, we explicitly allocate more control authority to prevent the robot from assisting towards the wrong goal.

We express each sub-policy towards a goal as a conditional probability distribution $ \pi(a|s,g)$ of taking an action $a$ at state $s$ for a goal $g$. 
 \begin{equation}
\pi(a|s) = \Sigma_{g \in \mathcal{G}}  \pi(a|s,g) p(g)
\end{equation}
The marginalized probability $\pi(a|s)$ is a mixture model 
which is a weighted sum over the conditional probabilities $ \pi(a|s,g)$ with probability $p(g)$, 
for a discrete set of goals $g \in \mathcal{G}$.
The decision points are identified
by reasoning on the modality of the distribution $\pi(a|s)$.
A single mode corresponds to having no decision
point, while multiple modes indicate a decision point.

\subsection{Directional Policies on the Plane}

Consider a 1-dimensional action case,
where the action is an angle. Each sub-policy $\pi(a|s,g)$ can be modeled as a von Mises distribution, as shown in Figure~\ref{fig:mixture_distributions}.
Von Mises distributions express a probability distribution
along all \textit{directions} in the plane, which is a continuous distribution on a unit circle~\cite{gumbel1953circular}.
The general von Mises probability density function is expressed as:
\begin{equation}
f(x|\;\mu, \kappa) = \frac{1}{2\pi I_0(\kappa)}\exp^{\kappa \cos(x-\mu)}
\end{equation}
where the parameters mode $\mu$ and dispersion $1 / \kappa$ are comparable to the mean $\mu$ and the variance $\sigma^{2}$ 
in the normal distribution, and $I_0(\kappa)$ is the modified Bessel function of order 0. 
In higher dimensions, it takes the form of a von Mises-Fisher distribution over a n-sphere. 

For each sub-policy, the mode $\mu$ of the distribution shows the direction of the action with the highest probability 
and the dispersion $\kappa$ is related to the scores from the goal prediction.

The resulting $\pi(a|s)$ is a Finite von Mises Mixture model (FVMM).
When the robot is near an obstacle or approaching to grab an object (situations where robot assistance is needed), 
all sub-policies are similar (co-directed) or one sub-policy dominates others (points 1 \& 2 in Figure~\ref{subfig-2:env}). 
The mixture distribution $\pi(a|s)$ is unimodal, as shown in Figures~\ref{subfig-2:p21}, \ref{subfig-2:p44}, 
and indicates the condition where the robot should take more control authority.

When the robot is near a decision point, 
e.g. approaching an obstacle or in between two goal objects (points 3 \& 4 in Figure~\ref{subfig-2:env}), 
the sub-policies disagree.
The resulting mixture distribution becomes multimodal 
with modes in the possible directions that the robot can take, 
as shown in Figures~\ref{subfig-3:p57}, \ref{subfig-3:p63}. 
Here, we allocate more control authority to the user.

\begin{algorithm}[h!]
  \DontPrintSemicolon
  \SetAlgoNoLine
   \KwIn{$s_t, a_t^{S}, g^*$}
   \KwOut{$r_t$}
   \SetKwFunction{FMain}{Reward}
   \SetKwProg{Fn}{Function}{:}{}
   \Fn{\FMain{$s_t, a_t^{S}, g^*$}}{
     Compute $R_{env}$ \;
    Means $\mathcal{M}=\{a^R_{t,g}\}^G_{g=1}$ \;
    Dispersions $\mathcal{K}=\{b_{t,g}\}^G_{g=1}$\; 
     Construct FVMM $\pi(a_t | s_t)=\frac{1}{2\pi I_0(\kappa)}\exp^{\kappa \cos(a_t-\mu)}$\;
     \If{multimodal}{
     $R_{agree}= -||a^{H}_t - a^{S}_t||^2$\;
     }
     \Else{$R_{agree}= -||a^{R}_{t,g^*} - a^{S}_t||^2$}
     $R_{speed}=-\big|\;||a^{H}_t|| - ||a^{S}_t||\;\big|$\;
     $R_{agree} \leftarrow R_{agree} + R_{speed}$\;
    $r_t  \leftarrow R_{agree}+R_{env}$\;
      \textbf{return} $r_t$\;
   }
   \textbf{End Function}\;
   \caption{Reward Function}
   \label{alg:reward}
  \end{algorithm}
\subsection{Reward Function using Disagreement}
The overall algorithm for computing the reward is summarized in Algorithm~\ref{alg:reward}.
The reward function consists of a reward term based on the agreement of the policies $R_{agree}$ 
and $R_{env}$ to penalize or reward certain actions while interacting with the environment. 
\begin{equation}
R(s,a,s') = R_{agree}+R_{env}
\end{equation}
$R_{env}$ includes negative reward when the robot is in collision with the obstacle (-10) or the workspace boundary (-2) and positive reward when the gripper reaches the goal object (+10).  
$R_{agree}$ is provided as a form of implicit feedback, by penalizing the disagreement between agent policies depending on the modality of the FVMM. When it is multimodal, $R_{agree}$ penalizes the L2 norm between the human agent $a^H$ and the arbitrated action $a^S$. When the FVMM is unimodal, $R_{agree}$ penalizes the L2 norm between the robot action $a^R_{g^*}$ and the arbitrated action $a^S$. The actions are normalized prior to computing the L2 norm and $a^R_{g^*}$ indicates the sub-policy for the true goal $g^*$ that the user intended, which is accessible in hindsight once the episode terminates. $R_{agree}$ is additionally subtracted with the absolute difference between the norms of $a^H_t$ and $a^S_t$ (denoted as $R_{speed}$) to match the policy's speed to that of the human's. 

There can be multiple ways to determine the modality of the FVMM. 
In the algorithm, we sample a number of points from the FVMM and examine whether a peak value exists 
that exceeds a threshold. 
When the FVMM is unimodal, there is one peak that exceeds the threshold due to its narrow dispersion. 
There are no peaks detected that exceeds the threshold when the FVMM is multimodal. 

%


\section{Experimental Setup}
\label{sec:experiment_setup}
We provide experimental details on training the model and evaluate the performance of our shared control agent. 
We hypothesize the following:
\begin{itemize}
\item The shared control agent can learn to flexibly allocate more control authority to the human when the prediction is unclear and provide more assistance when the goal is clear. 
\item The custom reward function enables faster convergence when training 
and it enables safe and accurate execution towards the user's intended goal.
\end{itemize}

\subsection{Environment Details}
We consider a teleoperation task where the human agent is controlling the end-effector 
of a robot manipulator to reach and grab the goal object while avoiding an obstacle. 
The simulated environment consists of a table with objects 
and a Baxter robot (see Figure \ref{fig:workflow}). 
The robot can manipulate its end-effector on a parallel plane 
above the 50cm$\times$50cm table workspace. 
Graspable objects are placed at the end of the workspace, 
and the obstacle is randomly located near the middle of the workspace. 
The gripper position is initialized to start behind the obstacle 
so that the robot arm encounters the obstacle when reaching for the goal object.
Physical collisions are not included in the simulated environment.

\subsection{Obtaining Robot Sub-policies}
\label{subsection:Subpolicies}
We train a goal-conditioned neural network policy $\pi_g$ by imitating
optimized robot grasping trajectories.
We obtain multiple sub-policies by pre-conditioning the policy on different goals.

The state and action space of our robot policy is solely defined in task space. 
Consequently, the network must learn to output end-effector actions 
that avoid collision with the whole arm.
We collect a dataset of 24K trajectories by computing pick-and-place motions 
with random starting positions and environment configurations 
using a Rapidly exploring Random Tree (RRT) 
followed by a Gauss-Newton trajectory optimizer~\cite{mainprice2016warping} algorithm 
that considers the full-arm kinematics and the 3D workspace.
In the training phase, we only consider the end-effector state
(i.e., position and velocity).

Each sub-policy $\pi_g: \tilde{s}_t \mapsto a^R_g$ maps
the state of the system
and outputs an action towards the goal.
The state $\tilde{s}_t$ is a concatenation of robot and environment states consisting of: 
end-effector position $p_{gripper}$, vector components of current gripper rotation in z-axis 
\begin{equation}
\mathbf{v}_{rot_Z}=[\cos\varphi , \sin\varphi]^T
\end{equation} 
obstacle position $p_{obstacle}$. 
In our experiment, we constrict our environment to a 2-dim plane over the workspace.

\subsection{Predicting User Intent}

In the experiments, we consider
a simple intent inference model
proposed in \cite{dragan2013policy}, based on
distance and direction towards the goal.

The score $b_g$ describes how likely the object is the goal object. 
Here, we use the posterior probability value for each object. 
The goal object $g^*$ can be chosen as the object that maximizes the posterior probability 
given a history of observed features $\xi_{S\rightarrow U}$~\cite{dragan2013policy, jain2019probabilistic}. 
\begin{equation}
g^* = \argmax_{g \in \mathcal{G}} P(g|\xi_{S\rightarrow U})
\end{equation}  
We follow Dragan and Srinivasa~\cite{dragan2013policy} to compute $P(g|\xi_{S\rightarrow U})$
 by using sum of squared velocity magnitudes as the cost function.

\begin{figure*}[t]
\centering\
  \begin{subfigure}{0.24\textwidth}
  \centering
  \includegraphics[clip,trim={0.2cm 0.2cm 0cm 0.7cm},width=1.1\textwidth]{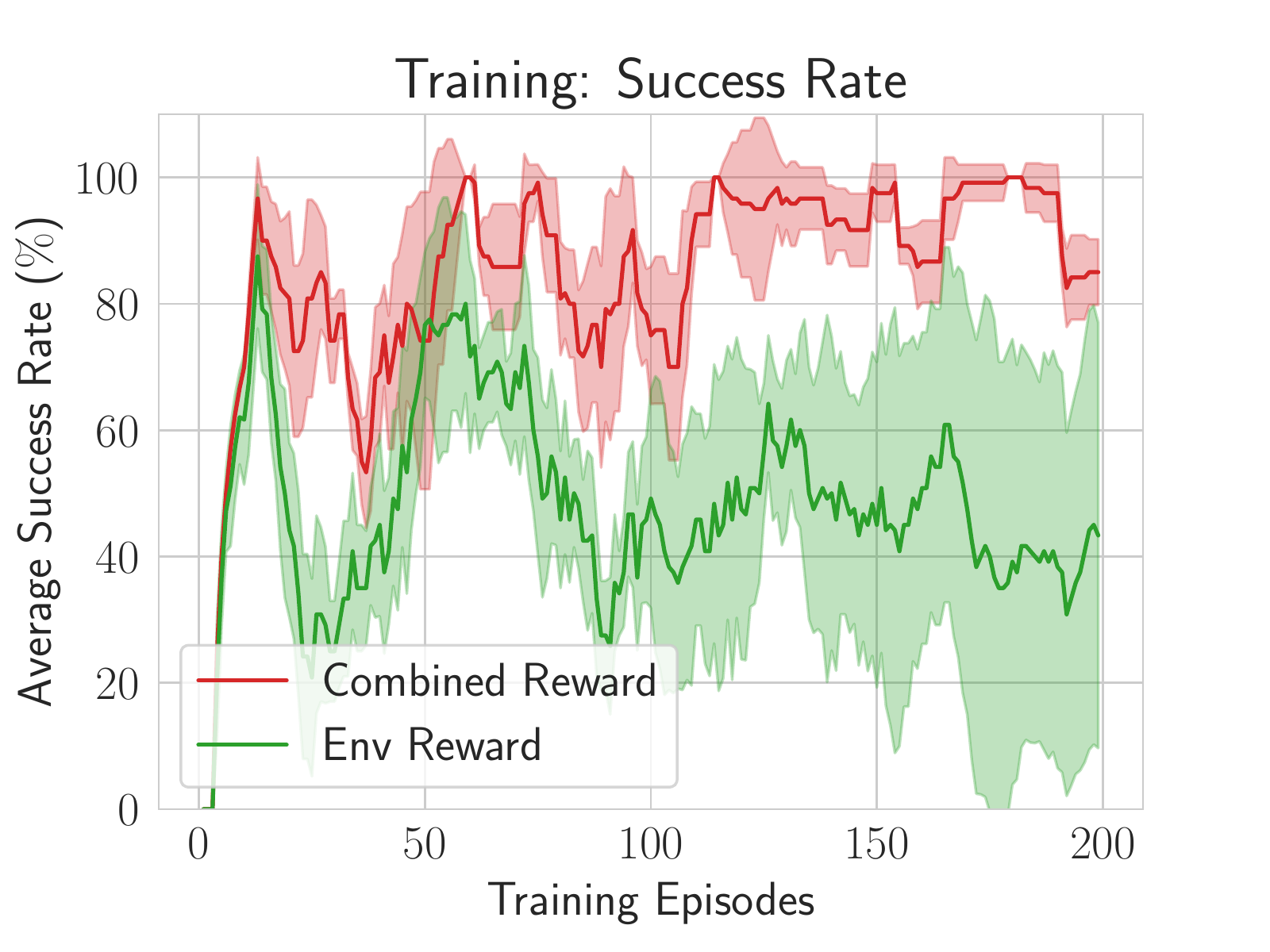}
  \label{subfig:grab}
  \end{subfigure}
  \hfill
  \begin{subfigure}{0.24\textwidth}
  \centering
  \includegraphics[clip,trim={0.2cm 0.2cm 0cm 0.7cm},width=1.1\textwidth]{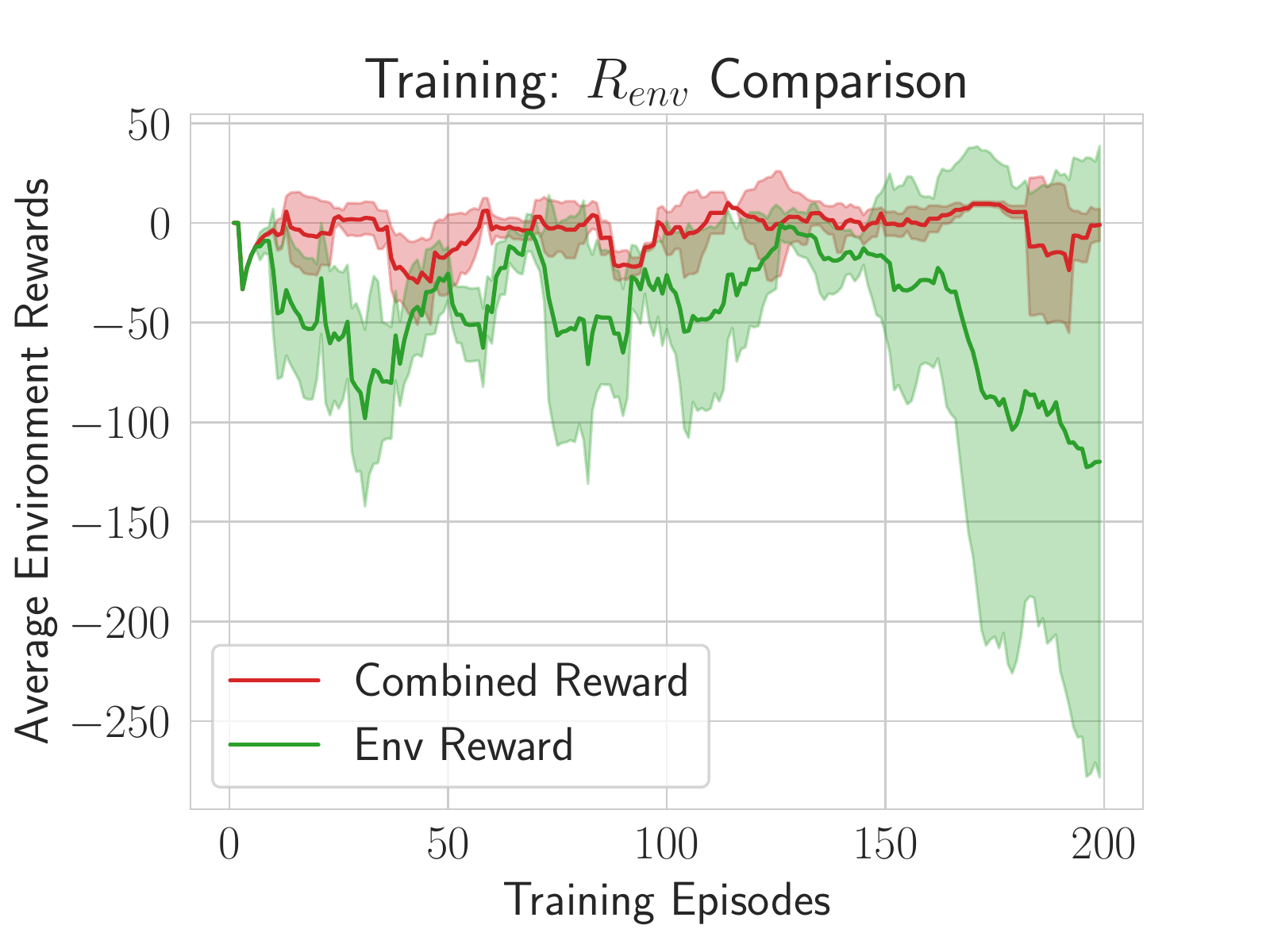}
  \label{subfig:proximity}
  \end{subfigure}
  \hfill
  \begin{subfigure}{0.24\textwidth}  
  \centering  
  \includegraphics[clip,trim={0.7cm 0cm 0cm 0.7cm},width=1.1\textwidth]{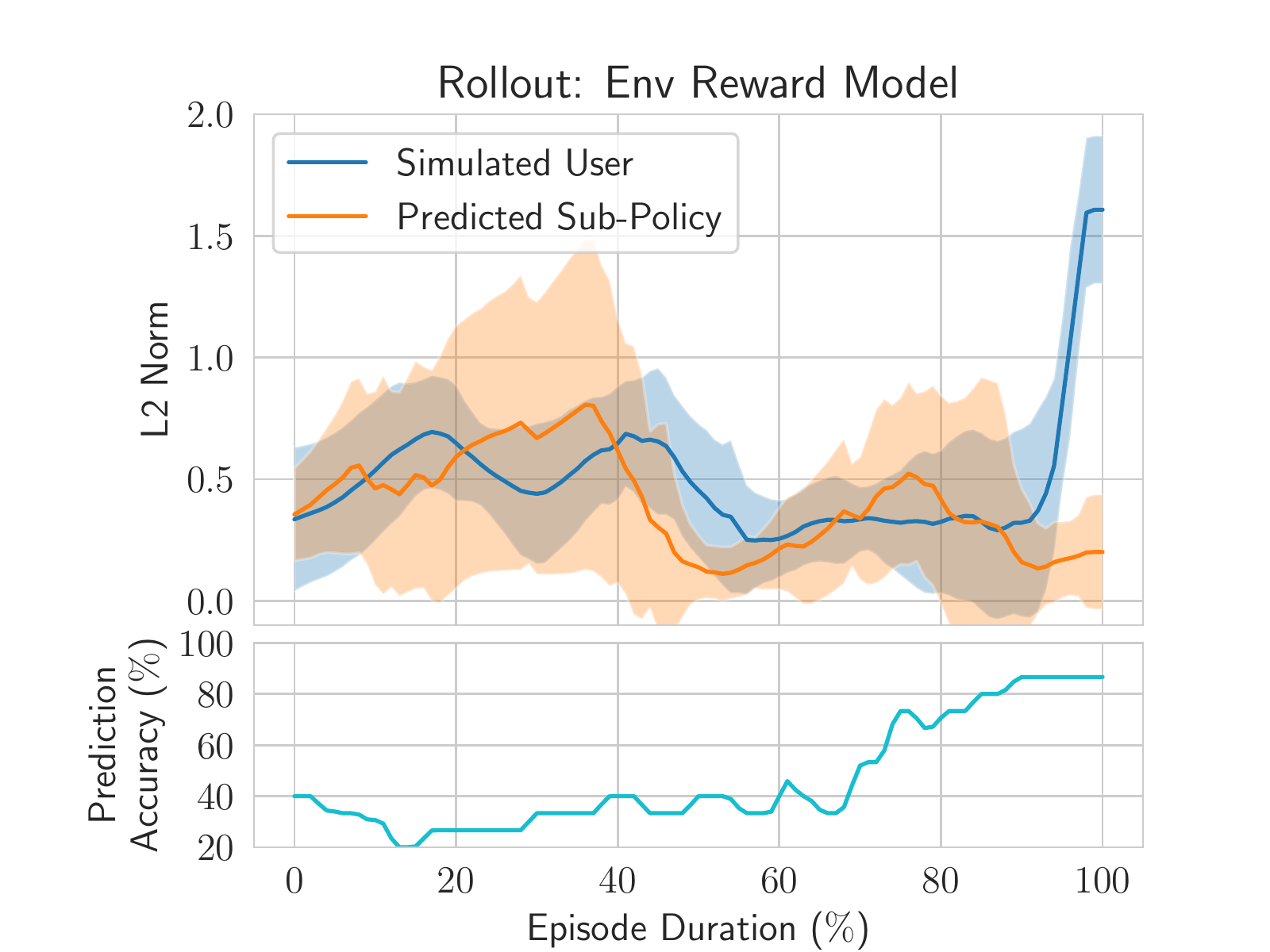}
  \label{subfig:rollout_env}
  \end{subfigure}
  \hfill
  \begin{subfigure}{0.24\textwidth}
  \centering
  \includegraphics[clip,trim={0.7cm 0cm 0cm 0.7cm},width=1.1\textwidth]{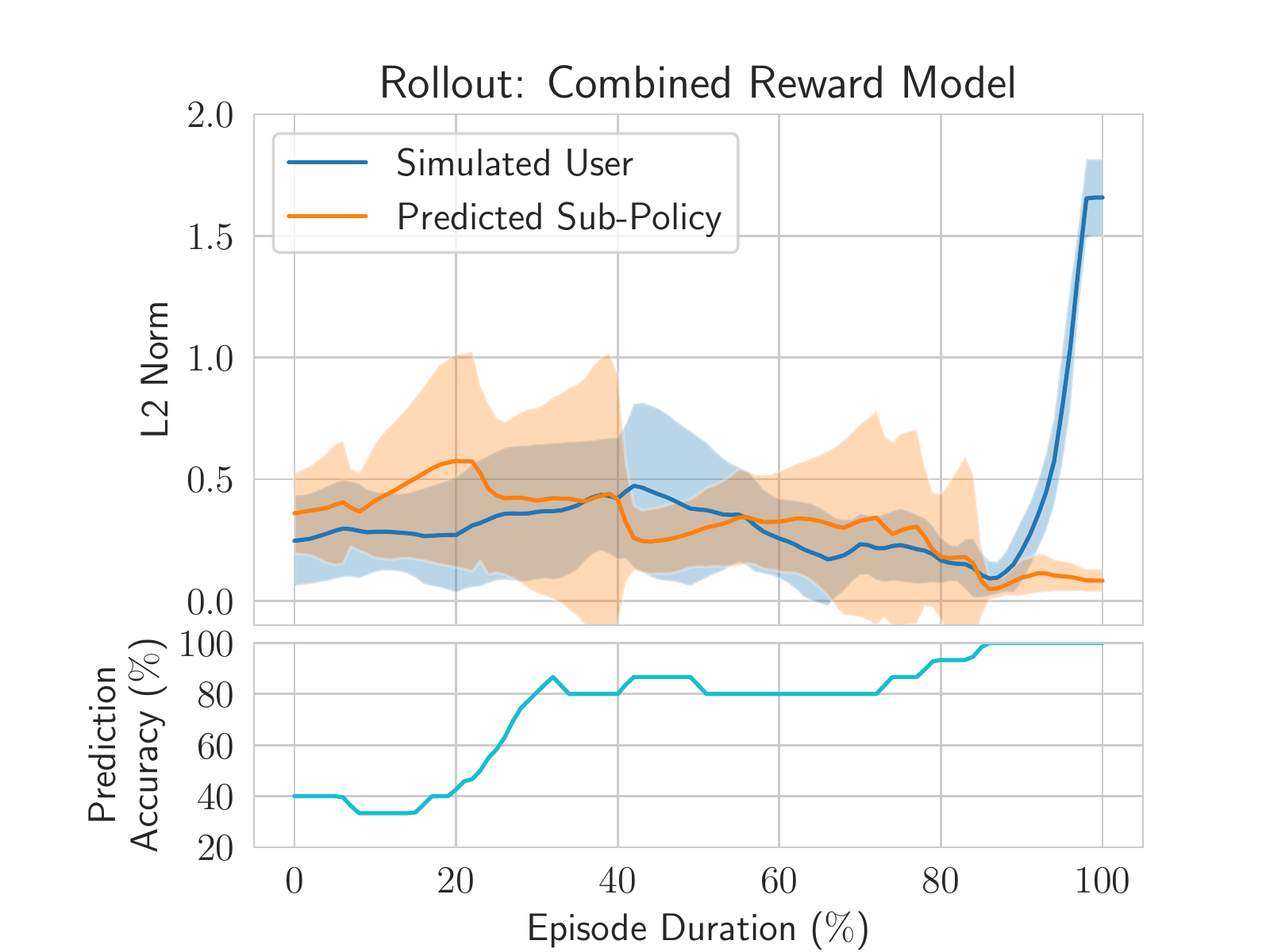}
  \label{subfig:rollout}
  \end{subfigure}
  \vspace{-0.5cm}
  \caption{(First two plots): Comparison between the policies with different reward functions during training, 
  each averaged across 12 trained models and smoothed using a moving average (window size: 10 episodes).\\  
  (Last two plots): Demonstrations of the trained models, each averaged across 15 random setting episodes.
  The blue line indicates the L2 norm between the human action and the arbitrated action 
  and the orange line indicates the L2 norm between the action from the sub-policy with the highest score 
  (predicted sub-policy) and the arbitrated action.}\label{fig:results}
  \vspace{-0.2cm}
\end{figure*}

\subsection{Simulating Human Users}
In the experiments, we simulate human policies to replace the interaction with the reinforcement learning agent during training and evaluation:
\begin{itemize}
\item \textit{Noisy} user as a sub-optimal noisy policy
\item \textit{Straight} user executes a policy that points directly to the goal
\item \textit{Biased} user misperceives the goal location due to imprecise perception (e.g. perceiving the goal closer than it actually is)
\end{itemize}
Both \textit{Noisy} and \textit{Biased} users make use of the policy described in Section~\ref{subsection:Subpolicies}. The \textit{Noisy} user adds random noise at each time step to the policy, whereas the \textit{Biased} user adds a random offset to the goal position.

\subsection{Training Details}
The DDPG arbitration module takes as input a 17-dimensional $s_t$ including
user action, all sub-policies and scores, and environment states (distance to goals, gripper position, and distance to the obstacle). 
$s_t$ excluding the user action is passed through three dense layers of 32 units 
followed by a layer of two units. We refer to this part of the network as the ``head''. 
The head is concatenated with the user action 
and passed through three dense layers of 16 units and outputs a continuous two-dimensional action, 
which represents the arbitrated action. This is referred to as the actor network.

The critic network uses the same head structure. 
The user action and the arbitrated action are concatenated with the output of the critic network's head 
and connected to two dense layers of 128 units. 
The output of the critic network is a one-dimensional value 
that estimates the expected return after taking an action at a state.

The head is pretrained such that it learns to output the sub-policy with the highest score 
given all sub-policies, scores, and the environment information.
This pretrained head is substituted in both actor, critic networks with frozen weights. 
The remaining layers of the arbitration module are also pretrained, 
such that the arbitration module eventually outputs the sub-policy 
with the highest score after the head network is concatenated with the user action. 
This way, the arbitration module is initialized to give full control to the robot 
and lowers its arbitration level as it interacts with the user. 

This pretraining step was essential to lower
the sample complexity of the algorithm. 
Without pretraining, the user must interact with random arbitrated actions
which would make the user fight for control authority during training. 
In addition, the random trajectory that the robot creates from random arbitrated actions makes it even difficult to predict the goal 
and it exacerbates the training by generating noisy prediction scores.

\section{Results}
\label{sec:results}
\subsection{Effects of Penalizing Disagreement during Training}
We aim to see whether our reward function is beneficial during training 
and can lead to successfully learning an arbitration policy.

As shown in the first plot in Figure~\ref{fig:results}, 
the combined reward model ($R_{agree}+R_{env}$) converged to a high success rate ($>90\%$) 
after 110 training episodes, whereas the success rate of the environment reward model ($R_{env}$) decreased over training.
Taking into account that our goal prediction model is not perfect, 
we can assume that the combined reward model learned to assign more control authority to the user 
in the decision points where the FVMM is multimodal, 
rather than solely relying on the prediction and assisting towards the goal with the highest score.

When comparing the $R_{env}$ term for both models (see Figure~\ref{fig:results} second plot), 
the combined reward model outperformed the environment reward model by maintaining a stable $R_{env}$. 
$R_{env}$ is penalized at each time step when there is a collision with the obstacle or goes out of bounds 
and rewarded once it reaches the goal object. 
The combined reward model assured safer teleoperation by having fewer collisions and remaining inside the workspace boundary. 
It learned to avoid obstacles by allocating more control authority to the robot near the obstacle, 
where the FVMM is unimodal since all sub-policies agree by pointing outwards away from the obstacle.

\subsection{Rolling Out the Trained Models}
We demonstrated rollouts using the trained models to see how the arbitration policy behaves during an episode.
We selected the trained models with the best return and demonstrated 15 randomized episodes. 
The \textit{Straight} user mode was used as the human policy for the last two plots in Figure~\ref{fig:results}.

As desired, the combined reward policy flexibly allocates the control authority depending on the situation.
In the beginning of the episode (duration $<30\%$), the arbitrated policy was closer to the human policy. 
This is when the robot arm approaches the first decision point and should decide 
which way to get around the obstacle with low prediction accuracy. 
In the middle of the episode when the robot gripper must get around the obstacle (duration $40\%\sim 60\%$), 
the arbitrated policy is closer to the predicted sub-policy, which shows that the robot takes more control near the obstacle. 
Towards the end of the episode (duration $60\%\sim80\%$) when the robot gripper encounters the second decision point to decide the goal object, 
the arbitrated policy is closer to the human policy, and the human gains more control authority. 
At the end of the episode, the arbitrated policy is closer to the predicted sub-policy, 
which provides more assistance when grabbing the object.

This effect was less prominent in the plot using the environment reward policy.
In addition, the L2 norms for the environment reward were comparably higher 
than that of the combined reward in the first half of the episode. 
The arbitrated policy was neither close to both human and predicted sub-policy, 
meaning that the arbitrated policy generated policies that do not comply with both policies.

\subsection{Comparison between Different Simulated Users}
Table~\ref{tab:table1} shows the result of demonstrations 
from the learned policies for all user modes: 
\textit{Noisy} user with two different noise sizes, \textit{Straight} user, and \textit{Biased} user. 
No assistance was applied in the direct control method.

Although the combined reward policy did not show the shortest travel distance among other assistance methods,
it achieved more successful episodes and fewer collisions for all users than the environment reward policy. 
A compromise in the travel distance would have been necessary to avoid the obstacle from a safe distance. 
It is clearly shown from the \textit{Straight} user 
that the combined reward policy had augmented the straight path to avoid the obstacle. 

With the \textit{Noisy} user, 
the combined reward policy showed a shorter travel distance than the direct control method, 
implying that the combined policy counterbalanced the noise. 
The combined reward policy was also beneficial for the \textit{Biased} user,
correcting the imprecise user commands to grab the goal object. 

Overall results show that the $R_{agree}$ term in the combined reward policy 
leads to successfully learning a policy that accurately assists towards the user's intended goal 
while maintaining safety. 

\begin{table}[h]
\setlength\tabcolsep{2pt} 

  \begin{center}
    \caption{Rollouts with simulated users over 15 episodes}
    \label{tab:table1}
    \begin{tabular}{c|c|c|c|c|c}
      \toprule 
      \textbf{User} &\textbf{Assistance} & \textbf{Success} & \textbf{Travel Dist.} & \textbf{Collisions} \\
      \textbf{Mode}& \textbf{Method} &  & (cm) & \\
      \midrule 
      \multirow{3}{*}{Noisy 0.5}&Combined Reward & \textbf{14/15} & 53.68 $\pm$ 3.17 & \textbf{0/15}\\
      &Environment Reward & 12/15 & \textbf{53.06 $\pm$ 2.64} & 2/15\\
      &Direct Control & \textbf{14/15} & 59.12 $\pm$ 16.09 & 1/15\\
      \midrule 
      \multirow{3}{*}{Noisy 1.0}&Combined Reward & \textbf{14/15} & 59.41 $\pm$ 18.70 & \textbf{0/15}\\
      &Environment Reward & 12/15 & \textbf{53.99 $\pm$ 4.06} & 2/15\\
      &Direct Control &  \textbf{14/15} & 71.16 $\pm$ 13.23 & \textbf{0/15}\\
      \midrule 
      \multirow{3}{*}{Straight}&Combined Reward & \textbf{13/15} & 51.63 $\pm$ 1.65 & \textbf{2/15}\\
      &Environment Reward & 10/15 & 54.02 $\pm$ 2.38 & 9/15 \\
      &Direct Control & 11/15 & \textbf{49.19 $\pm$ 2.17} & 7/15\\
      \midrule 
      \multirow{3}{*}{Biased}&Combined Reward & \textbf{8/15} & \textbf{54.60 $\pm$ 11.81} & \textbf{0/15}\\
      &Environment Reward & 6/15 & 62.54 $\pm$ 21.46 & 1/15\\
      &Direct Control & 3/15 & 65.27 $\pm$ 22.21 & \textbf{0/15} \\
      \bottomrule 
    \end{tabular}
  \end{center}

\end{table}

\section{Conclusion}
\label{sec:conclusions}
We proposed a framework to learn an arbitrated policy for shared control 
using DDPG that can dynamically hand over control authority to the user at a decision point. 
We identified the decision points by looking at the diversity of the sub-policies and 
constructing a Finite von Mises Mixture model to observe the modality of the distribution. 
Experiment results indicate that incorporating the implicit feedback 
allows the agent to effectively learn when to hand over control authority 
while maintaining safe and accurate teleoperation despite an imperfect goal prediction system.
A limitation of our work is the lack of user interaction, which we will investigate as future work.


\section*{ACKNOWLEDGMENT}

This work is partially funded by the research alliance ``System Mensch''.
The authors thank the International Max Planck Research School for Intelligent Systems (IMPRS-IS) for supporting Yoojin Oh.


\bibliographystyle{IEEEtran}
\bibliography{bibliography}{}

\begin{thebibliography}{10}
\providecommand{\url}[1]{#1}
\csname url@rmstyle\endcsname
\providecommand{\newblock}{\relax}
\providecommand{\bibinfo}[2]{#2}
\providecommand\BIBentrySTDinterwordspacing{\spaceskip=0pt\relax}
\providecommand\BIBentryALTinterwordstretchfactor{4}
\providecommand\BIBentryALTinterwordspacing{\spaceskip=\fontdimen2\font plus
\BIBentryALTinterwordstretchfactor\fontdimen3\font minus
  \fontdimen4\font\relax}
\providecommand\BIBforeignlanguage[2]{{%
\expandafter\ifx\csname l@#1\endcsname\relax
\typeout{** WARNING: IEEEtran.bst: No hyphenation pattern has been}%
\typeout{** loaded for the language `#1'. Using the pattern for}%
\typeout{** the default language instead.}%
\else
\language=\csname l@#1\endcsname
\fi
#2}}

\bibitem{phillips2016autonomy}
C.~Phillips-Grafflin, \emph{et~al.}, ``From autonomy to cooperative traded
  control of humanoid manipulation tasks with unreliable communication,''
  \emph{Journal of Intelligent \& Robotic Systems}, vol.~82, no. 3-4, pp.
  341--361, 2016.

\bibitem{johns2016exploring}
M.~Johns, \emph{et~al.}, ``Exploring shared control in automated driving,''
  \emph{ACM/IEEE Int. Conf. on Human-Robot Interaction (HRI)}, 2016.

\bibitem{muelling2017autonomy}
K.~Muelling, \emph{et~al.}, ``Autonomy infused teleoperation with application
  to brain computer interface controlled manipulation,'' \emph{Autonomous
  Robots}, vol.~41, no.~6, pp. 1401--1422, 2017.

\bibitem{goil2013using}
A.~Goil, \emph{et~al.}, ``Using machine learning to blend human and robot
  controls for assisted wheelchair navigation,'' in \emph{2013 IEEE 13th
  International Conference on Rehabilitation Robotics (ICORR)}, 2013.

\bibitem{lillicrap2015continuous}
T.~P. Lillicrap, \emph{et~al.}, ``Continuous control with deep reinforcement
  learning,'' \emph{arXiv preprint arXiv:1509.02971}, 2015.

\bibitem{dragan2013policy}
A.~D. Dragan and S.~S. Srinivasa, ``A policy-blending formalism for shared
  control,'' \emph{The International Journal of Robotics Research}, vol.~32,
  no.~7, pp. 790--805, 2013.

\bibitem{allaban2018blended}
A.~A. Allaban, \emph{et~al.}, ``A blended human-robot shared control framework
  to handle drift and latency,'' \emph{arXiv preprint arXiv:1811.09382}, 2018.

\bibitem{schultz2017goal}
C.~Schultz, \emph{et~al.}, ``Goal-predictive robotic teleoperation from noisy
  sensors,'' \emph{IEEE Int. Conf. Robotics And Automation (ICRA)}, 2017.

\bibitem{gopinath2016human}
D.~Gopinath, \emph{et~al.}, ``Human-in-the-loop optimization of shared autonomy
  in assistive robotics,'' \emph{IEEE Robotics and Automation Letters}, vol.~2,
  no.~1, pp. 247--254, 2016.

\bibitem{kim2011autonomy}
D.-J. Kim, \emph{et~al.}, ``How autonomy impacts performance and satisfaction:
  Results from a study with spinal cord injured subjects using an assistive
  robot,'' \emph{IEEE Transactions on Systems, Man, and Cybernetics-Part A:
  Systems and Humans}, vol.~42, no.~1, pp. 2--14, 2011.

\bibitem{Javdani:2018bt}
S.~Javdani, \emph{et~al.}, ``Shared autonomy via hindsight optimization for
  teleoperation and teaming,'' \emph{The International Journal of Robotics
  Research}, vol.~37, no.~7, pp. 717--742, 2018.

\bibitem{broad2018operation}
A.~Broad, \emph{et~al.}, ``Operation and imitation under safety-aware shared
  control,'' in \emph{Workshop on the Algorithmic Foundations of Robotics},
  2018.

\bibitem{broad2019highly}
------, ``Highly parallelized data-driven mpc for minimal intervention shared
  control,'' \emph{Robotics: Science and Systems (RSS)}, 2019.

\bibitem{0h:20}
Y.~Oh, \emph{et~al.}, ``Natural gradient shared control,'' \emph{IEEE Int.
  Symp. on Robot and Human Interactive Communication (RO-MAN)}, 2020.

\bibitem{mnih2015human}
V.~Mnih, \emph{et~al.}, ``Human-level control through deep reinforcement
  learning,'' \emph{Nature}, vol. 518, no. 7540, pp. 529--533, 2015.

\bibitem{silver2016mastering}
D.~Silver, \emph{et~al.}, ``Mastering the game of go with deep neural networks
  and tree search,'' \emph{Nature}, vol. 529, no. 7587, pp. 484--489, 2016.

\bibitem{xu2015reinforcement}
W.~Xu, \emph{et~al.}, ``Reinforcement learning-based shared control for
  walking-aid robot and its experimental verification,'' \emph{Advanced
  Robotics}, vol.~29, no.~22, pp. 1463--1481, 2015.

\bibitem{gaoxt2}
J.~Gao, \emph{et~al.}, ``Xt2: Training an x-to-text typing interface with
  online learning from implicit feedback,'' \emph{International Conference on
  Learning Representations (ICLR)}, 2021.

\bibitem{reddy2018shared}
S.~Reddy, \emph{et~al.}, ``Shared autonomy via deep reinforcement learning,''
  \emph{arXiv preprint arXiv:1802.01744}, 2018.

\bibitem{schaff2020residual}
C.~Schaff and M.~R. Walter, ``Residual policy learning for shared autonomy,''
  \emph{arXiv preprint arXiv:2004.05097}, 2020.

\bibitem{fernandez2021deep}
F.~C. Fernandez and W.~Caarls, ``Deep reinforcement learning for haptic shared
  control in unknown tasks,'' \emph{arXiv preprint arXiv:2101.06227}, 2021.

\bibitem{du2020ave}
Y.~Du, \emph{et~al.}, ``Ave: Assistance via empowerment,'' \emph{Neural
  Information Processing Systems (NeurIPS)}, 2020.

\bibitem{jain2019probabilistic}
S.~Jain and B.~Argall, ``Probabilistic human intent recognition for shared
  autonomy in assistive robotics,'' \emph{ACM Transactions on Human-Robot
  Interaction (THRI)}, vol.~9, no.~1, pp. 1--23, 2019.

\bibitem{gumbel1953circular}
E.~Gumbel, \emph{et~al.}, ``The circular normal distribution: Theory and
  tables,'' \emph{Journal of the American Statistical Association}, vol.~48,
  no. 261, pp. 131--152, 1953.

\bibitem{mainprice2016warping}
J.~Mainprice, \emph{et~al.}, ``Warping the workspace geometry with electric
  potentials for motion optimization of manipulation tasks,'' \emph{IEEE/RSJ
  Int. Conf. on Intel. Robots And Systems (IROS)}, 2016.

\end{thebibliography}

\end{document}